\documentclass[conference]{IEEEtran}
\IEEEoverridecommandlockouts
\usepackage{cite}
\usepackage{amsmath,amssymb,amsfonts}
\usepackage{algorithmic}
\usepackage{graphicx}
\usepackage{textcomp}
\usepackage{xcolor}
\usepackage{wrapfig}
\usepackage[a4paper, total={184mm,239mm}]{geometry}
\def\BibTeX{{\rm B\kern-.05em{\sc i\kern-.025em b}\kern-.08em
    T\kern-.1667em\lower.7ex\hbox{E}\kern-.125emX}}

\usepackage{hyperref}
\hypersetup{
    colorlinks=true,
    citecolor=blue,
    linkcolor=blue,
    urlcolor=blue,
}

\usepackage{amsmath}
\usepackage{cleveref}
\usepackage{multirow}
\usepackage{listings}
\usepackage{subcaption}
\usepackage{enumitem}

\usepackage{tikz}
\newcommand*\circled[1]{\tikz[baseline=(char.base)]{
            \node[shape=circle,draw,inner sep=0.5pt] (char) {#1};}}
            
\newcommand{\eg}{\mbox{{\em e.g.}}}
\newcommand{\ie}{\mbox{{\em i.e.}}}

\newcommand{\cf}{\mbox{{c.f.}}}
\newcommand{\bem}[1]{{\bf\em #1}}
\newcommand{\fixme}[1]{\textcolor{black}{#1}}

\newcommand{\goldmine}{\mbox{{\scshape GoldMine}}}
\newcommand{\gptt}{\mbox{{GPT-3.5}}}
\newcommand{\gptf}{\mbox{GPT-4o}}
\newcommand{\cllama}{\mbox{CodeLLaMa 2}}
\newcommand{\llama}{\mbox{LLaMa3-70B}}

\newcommand{\pname}{\mbox{{AssertionBench}}}
\newcommand{\allm}{\mbox{{AssertionLLM}}}

\definecolor{codegreen}{rgb}{0,0.6,0}
\definecolor{codegray}{rgb}{0.5,0.5,0.5}
\definecolor{codepurple}{rgb}{0.58,0,0.82}
\definecolor{backcolour}{rgb}{0.95,0.95,0.92}

\lstdefinestyle{mystyle}{
    backgroundcolor=\color{backcolour},   
    commentstyle=\color{codegreen},
    keywordstyle=\color{blue},
    numberstyle=\tiny\color{codegray},
    stringstyle=\color{codepurple},
    basicstyle=\ttfamily\footnotesize,
    breakatwhitespace=false,         
    breaklines=true,                 
    captionpos=b,                    
    keepspaces=true,                 
    numbers=left,                    
    numbersep=5pt,                  
    showspaces=false,                
    showstringspaces=false,
    showtabs=false,                  
    tabsize=2
}

\begin{document}

\title{Are LLMs Ready for Practical \\ Adoption for Assertion Generation?
}
\author{
\IEEEauthorblockN{Vaishnavi Pulavarthi, Deeksha Nandal}
\IEEEauthorblockA{\textit{Electrical and Computer Engg. Dept.} \\
\textit{University of Illinois Chicago}\\
Chicago, USA \\
\{vpulav2, dnanda6\}@uic.edu}
\and
\IEEEauthorblockN{Soham Dan}
\IEEEauthorblockA{\textit{Microsoft} \\
sohamdan@microsoft.com}
\and
\IEEEauthorblockN{Debjit Pal}
\IEEEauthorblockA{\textit{Electrical and Computer Engg. Dept.} \\
\textit{University of Illinois Chicago}\\
Chicago, USA \\
dpal2@uic.edu}
}

\maketitle

\begin{abstract}
Assertions have been the de facto collateral for simulation-based and formal verification of hardware designs for over a decade. The quality of hardware verification, \ie, detection and diagnosis of corner-case design bugs, is critically dependent on the quality of the assertions. With the onset of generative AI such as Transformers and Large-Language Models (LLMs), there has been a renewed interest in developing novel, effective, and scalable techniques of generating functional and security assertions from design source code. While there have been recent works that use commercial-of-the-shelf (COTS) LLMs for assertion generation, there is no comprehensive study in quantifying the effectiveness of LLMs in generating syntactically and semantically correct assertions. In this paper, we first discuss $\pname$ from our prior work, a comprehensive set of designs and assertions to quantify the goodness of a broad spectrum of COTS LLMs for the task of assertion generations from hardware design source code. Our key insight was that COTS LLMs are not yet ready for prime-time adoption for assertion generation as they generate a considerable fraction of syntactically and semantically incorrect assertions. Motivated by the insight, we propose $\allm$, a first of its kind LLM model, specifically fine-tuned for assertion generation. Our initial experimental results show that $\allm$ considerably improves the semantic and syntactic correctness of the generated assertions over COTS LLMs.

\end{abstract}

\section{Introduction}\label{sec:intro}

System-on-Chip (SoC) designs are crucial for many safety-critical computing applications, including vehicular systems, military, and industrial automation. SoCs often use sensitive and users' private data to perform numerous complex computations. It is crucial for our national and personal well-being to ensure that the SoCs are functionally correct, safe, and secure. 

\bem{Assertions} 
are mathematical encoding 
of desired design properties that should hold True for a design. 
Assertions are widely used for hardware design validation throughout its life cycle, 
\eg, pre-silicon formal verification and simulation-based verification, emulation, and often synthesized in a fabricated chip for post-silicon validation and in-field debug and diagnosis. In the past decade, assertion-based Verification (ABV)~\cite{witharana2022assertion} has 
emerged as the de facto standard to verify the security and functional correctness 
of hardware designs. 
However, crafting a succinct yet expressive set of assertions that capture subtle and important 
hardware design behaviors is a tedious and time-consuming task, 
requiring a considerable amount of human ingenuity. Too many assertions (i) can negatively affect 
verification performance with a prolonged verification closure and (ii) may require a substantial amount of on-chip resources, whereas too few assertions may result in insufficient design coverage causing corner case design bugs to escape to production and mass manufacturing. The ever increasing hardware design complexity and rapidly broadening target applications (\eg, deep learning, AI) have only worsened the problem. 
\bem{Consequently, developing an automated and scalable technique is crucial to rapidly generate a succinct set of hardware design properties targeting design functionality and security}.

A considerable amount of research has leveraged two different paradigms -- lightweight static analysis of design source code and formal verification 
~\cite{bensalem1996powerful, tiwari2004tacas, spin2004}, 
and data-driven statistical analysis, \eg, data mining~\cite{iodine2005, pinter2005hase, hekmatpour2005, inferno2009tcad, chang2015aspdac, chung2011iccd}. 
While 
static analysis can generalize 
and capture corner-case design behaviors, it suffers from prohibitive computational complexity limiting its scalability to industrial-scale designs. Alternatively, 
data-driven dynamic analysis can scale to large designs with a considerable amount of trace data due to its computational efficiency, however, it often generates spurious design properties due to the lack of design insights and domain context. 
More recently, researchers have proposed assertion generation techniques that combine static analysis and data-driven dynamic analysis~\cite{goldmine, iman2024artmine, harm2022tcad} and developed algorithms to induce ranking on such automatically generated assertions based on the subtle design behavior they capture~\cite{pal2020tcad}. However, all such techniques generate a large number of assertions, many of which are redundant and do not capture model-level or system-level design behaviors, and fail to scale to large industry-scale designs due to the algorithmic complexity of the underlying static analysis. Despite intense research across academia and industry over the last decade, there is a widening gap between assertion solutions and the industry's actual requirements in terms of assertion quality for complex hardware designs.  

With recent advances in deep-learning (DL) and generative AI models, especially Large-Language Models (LLMs), \eg, $\gptt$, $\gptf$, $\cllama$, and $\llama$, there is a renewed interest 
to harness the power of LLMs to tame the ever-widening gap. 
Most recent assertion generation approaches (\cf,~\Cref{sec:rel_work}) treat a LLM model as a \bem{black box} and use {\em prompt engineering} to iteratively refine the set of generated assertions. 
However, there is no in-depth study nor a dataset to evaluate the fit of 
different state-of-the-art (SOTA) LLM models 
for generating a succinct and correct set of assertions without a considerable amount of designer-developed prompts.

In this work, first, we discuss our prior work $\pname$ ~\cite{pulavarthi2025naacl}, the first comprehensive benchmark consisting of 100 curated hardware designs of varying complexity and their formally verified assertions to quantify the efficacy of SOTA and upcoming LLMs for 
assertion generation. 
Our primary focus is to quantify the quality 
of the generated assertions from SOTA LLMs 
learned on a set 
of labeled designs and their formally verified assertions. Our \bem{key insight} is that almost all SOTA LLMs generate a considerable fraction of syntactically and semantically incorrect assertions. \bem{Leveraging this insight, we develop $\allm$, a fine-tuned LLM model that can automatically generate substantially higher fraction (up to 25\%) of syntactically and semantically correct assertions from design source codes without any iterative inputs from the verification engineer}. We further outline several research \bem{challenges} and \bem{opportunities} that are worth pursuing to truly exploit the potential of generative AI for assertion generation. 


\section{Background 
}\label{sec:background}


\begin{figure}
\centering
\begin{lstlisting}[language=Verilog, 
                 %linewidth=2.8in, 
                 framexleftmargin=2pt,
                 framexrightmargin=2pt,
                 ]
  module arb2(clk, rst, req1, req2, gnt1, gnt2);
  input clk, rst, req1, re2;
  output gnt1, gnt2;
  reg gnt_, gnt1, gnt2;
  always @(posedge clk or posedge rst)
    if(rst)
       gnt_ <= 0;
    else
       gnt_ <= gnt1;
  always @(*)
    if (gnt_)
    begin
       gnt1 = req1 & req2;
       gnt2 = req2;
    end
    else
    begin
      gnt1 = req1;
      gnt2 = req2 & ~req1;
    end
  endmodule
\end{lstlisting}
\caption{{\bf A Verilog code for a 2-port Arbiter}~\cite{pal2020tcad}.}
\vspace{-4mm}
\label{fig:arbiter_code}
\end{figure}

\renewcommand{\arraystretch}{1.2}
\renewcommand{\columnsep}{10pt}
\begin{figure}
    \centering
    \resizebox{0.5\linewidth}{!}{
    \begin{tabular}{|cc|c|c|c|}
         \hline
         \multicolumn{3}{|c|}{}& \multicolumn{2}{|c|}{\bf Pre-condition}\\
         \cline{4-5}
         \multicolumn{3}{|c|}{} & {\bf Covered} & {\bf Unreachable}\\
         \hline
         \hline
         & & \multirow{2}{*}{\rotatebox[origin=c]{90}{{\centering \bf True}}} &  \multirow{2}{*}{\bf Valid} & \multirow{4}{*}{\centering \bf Vacuous}\\
         \multicolumn{2}{|c|}{\bf Post-} & 
         &  & \\
         \cline{3-4}
         \multicolumn{2}{|c|}{\bf Condition} & \multirow{2}{*}{\rotatebox[origin=c]{90}{{\centering \bf False}}} & \multirow{2}{*}{\bf CEX} & \\
         & & & & \\
         \hline
    \end{tabular}
    }
    \caption{{\bf Assertion status based on pre-condition and post-condition evaluation}. {\bf CEX}: Counter example.}
    \label{tab:assertion_eval}
    \vspace{-5mm}
\end{figure}

\subsection{Assertions: Syntax, Semantics, and Validity}\label{sec:assertions_syn_sem_val}
We consider a hardware design ${\cal D}$ in Verilog\footnote{We consider Verilog as the demonstration vehicle for this work, however, our work can naturally be extended to other hardware design languages, \eg, VHDL, SystemC, and other hardware description languages.} 
as a composition of a set of concurrent processes $P_i$, \eg, ({\tt always} and {\tt assign} blocks). Let $V$ be the set of design signals, 
${\cal I} \subset V$ be the set of input signals, ${\cal O} \subset V$ be the set of output signals, and ${\cal R} \subset V$ be the set of 
registers. 
\Cref{fig:arbiter_code} shows 
a Verilog design $\mathcal{D}$ of a 2-port Arbiter, consisting of 
two concurrent 
processes $P_1$ (line 6) and $P_2$ (line 11), and $V = \{clk, rst, req1, req2, gnt1, gnt2, gnt\_\}$, $\mathcal{I} = \{clk, rst, req1, req2\}$, $\mathcal{O} = \{gnt1, gnt2\}$, and $\mathcal{R} = \{gnt\_\}$.

An assertion is a temporal formula in LTL~\cite{ltl} of the format $P = \mathcal{G}(A \rightarrow C)$ where the antecedent $A$ is of the form $A = \bigwedge_{i = 0}^m {\cal X}^i(A^i)$ and consequent $C$ is of the form $C = \mathcal{X}^n(C^n)$, where $n \geq m$. Each $A_i$ ($C_i$) is a proposition and is a $(var, val)$ pair where $var \in V$ and $val \in \{0, 1\}$. $\mathcal{X}$ is called the next-cycle operator and $\mathcal{X}^i (i \geq 0)$ is equal to a delay of $i$ clock cycles $\underbrace{\mathcal{X}\mathcal{X}\ldots\mathcal{X}}_{i~\text{times}}$. Although SystemVerilog Assertion (SVA)~\cite{svmlrm} defines a rich set of grammar for assertions, we consider a restricted subset (sequential assertion) as captured by $P$. We say an assertion $P$ is {\bf True} (Valid) if $\mathcal{D} \models P$ (read as $\mathcal{D}$ models $P$), 
otherwise, the assertion is {\bf False}, \ie, $\mathcal{D} \not\models P$ and there exists a Boolean value assignment to a subset of design signals known as {\bf CEX} (counter-example) that shows a refutation of the assertion $P$ on $\mathcal{D}$. The implication operator $\rightarrow$ in $P$ are 
of two types, \bem{overlapped} and \bem{non-overlapped}. The {\em overlapped} implication operator ($\rightarrow$) implies {\em if there is a match on the antecedent $A$, then the consequent $C$ is evaluated in the same clock cycle} whereas the {\em non-overlapped} implication operator ($\Rightarrow$) implies {\em if there is a match on the antecedent $A$, then the consequent $C$ is evaluated in the next clock cycle}. 
In~\Cref{tab:assertion_eval}, we show the assertion evaluation status. Note $A \rightarrow C$ can be re-written as $\neg A \vee C$. Consequently, if pre-condtion $A$ is {\bf unreachable} (or False), then the assertion $P$ is vacuously True (\ie, $\neg False \vee C$). If pre-condition $A$ is True and post-condition $C$ is True as well, the assertion is reported to be Valid 
(\ie, $\mathcal{D} \models P$), otherwise, if the post-condition $C$ is False, then a counter example {\bf CEX} is generated. 

For the Arbiter 
of~\Cref{fig:arbiter_code}, consider 
assertions $P1: \mathcal{G}((req1 == 1 \wedge req2 == 0) \rightarrow (gnt1 == 1))$ and $P2: \mathcal{G}((req2 == 0 \wedge gnt\_ == 1) \wedge \mathcal{X}(req1 == 1) \rightarrow \mathcal{X}(gnt1 == 1))$. The assertion $P1$ 
evaluates 
{\bf True} if $req1$ is {\tt 1'b1} and $req2$ is {\tt 1'b0} at the current clock cycle, then $gnt1$ is {\tt 1'b1} in the current clock cycle. 
The assertion $P2$ 
evaluates 
{\bf True} if $req2$ is {\tt 1'b0} and $gnt\_$ is {\tt 1'b1} in the current cycle, $req1$ is {\tt 1'b1} in the next cycle, then $gnt1$ is {\tt 1'b1} in the cycle after (\ie, in the $2^{nd}$ cycle). Note that $P2$ can be re-written using the {\em non-overlapped} implication operator, 
$P2: \mathcal{G}((req2 == 0 \wedge gnt\_ == 1) \wedge \mathcal{X}(req1 == 1) \Rightarrow (gnt1 == 1))$ where the $\Rightarrow$ subsumes the $\mathcal{X}$ operator in 
the consequent. On 
discharging a proof for $P1$ and $P2$ using a formal property verification (FPV) engine\footnote{We use Cadence JasperGold~\cite{jaspergold}, however, any other FPV tool will work.}, 
we find $P1$ is a {\bf valid} assertion 
whereas $P2$ generates a {\bf CEX}. 

\subsection{Large-Language Models}\label{sec:llms}

Large-Language Models (LLMs) are an instance of generative AI built on top of encoder-decoder transformer architectures~\cite{attention2017nips}. LLMs can be classified primarily in three classes, (i) encoder-only LLMs~\cite{codebert2020emnlpf}, (ii) decoder-only LLMs~\cite{openai2024gpt4}, and (iii) encoder-decoder LLMs~\cite{t52021icse}. Encoder-only LLMs employ a bi-directional transformer during pre-training for each token to attend every other token, decoder-only LLMs employ unidirectional language modeling for each token to attend its predecessor tokens, and where tokens can only participate in previous tokens, and encoder-decoder LLMs employ denoising sequence-to-sequence pre-training objectives. The decoder-only LLM performs excellently in auto-regressive tasks such as code completion and generation. Since assertion generation is a special kind of code generation, in this work, we employ decoder-only LLMs, \eg, $\gptt$, $\gptf$, etc.

There are two distinct paradigms for LLM usage for different tasks -- (i) in-context learning (ICL), where a foundational LLM (\eg, $\gptf$) is seeded with a few examples of the desired task followed by deployment and (ii) fine tuning where a foundational LLM is trained with a small amount of high-quality downstream task-specific data to construct a task-specific LLM. In this work, we use ICL to evaluate the fitness of COTS LLMs for assertion generation and use finetuning to develop specialized LLMs for assertion generation tasks.



\section{$\pname$: Benchmark to quantify goodness of LLMs for assertion generation}\label{sec:benchmark}


\begin{figure}
    \centering
    \includegraphics[width=\columnwidth]{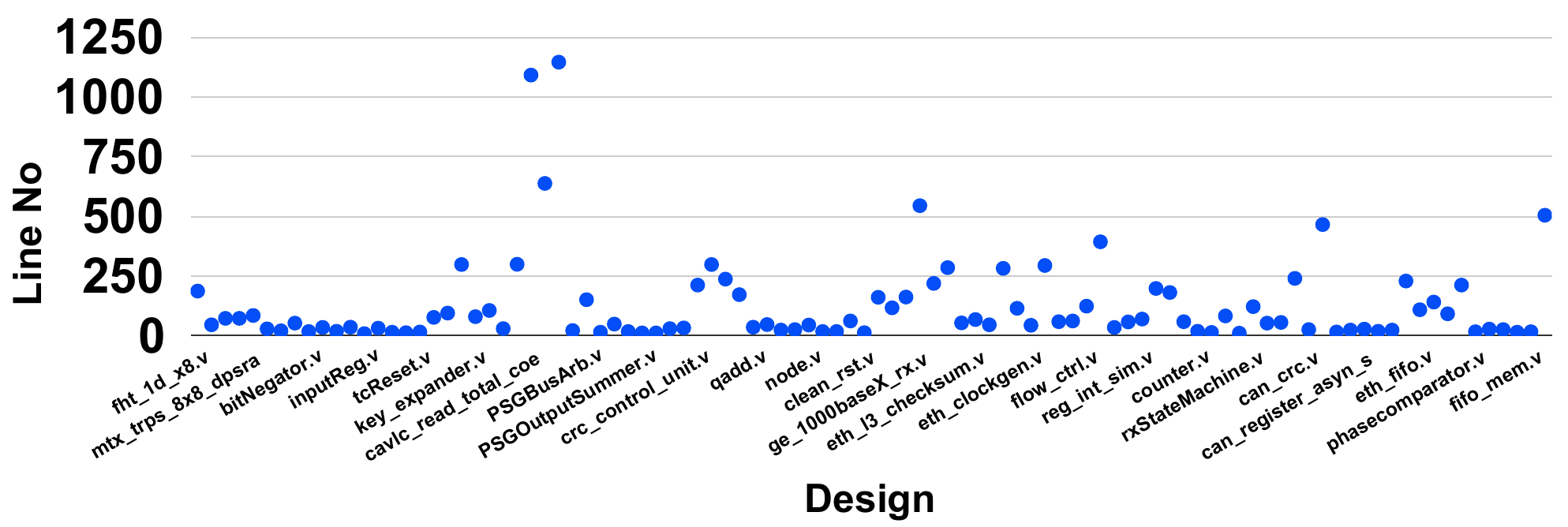}
    \vspace{-4mm}
    \caption{{\bf Design details in the test set in terms of the number of lines of code (excluding comments and blank lines)}.}
    \label{fig:design_deatails}
    \vspace{-5mm}
\end{figure}

\begin{figure}
    \centering
    \includegraphics[width=\columnwidth]{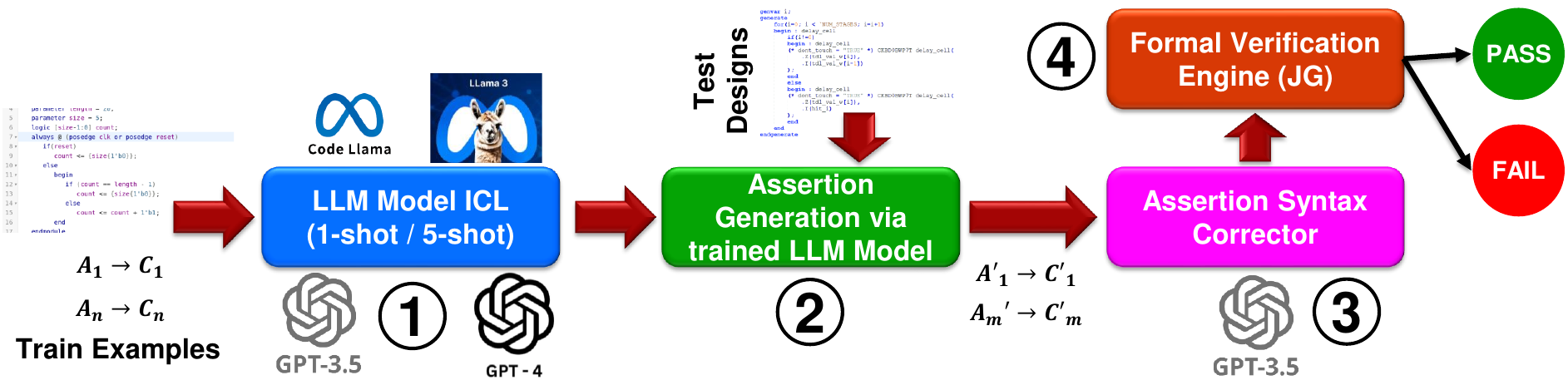}
    \vspace{-4mm}
    \caption{{\bf Framework to evaluate LLMs for assertion generation}~\cite{pulavarthi2025naacl}. {\bf JG}: 
    JasperGold Formal Property Verification Engine.}
    \label{fig:eval_framework}
    \vspace{-5mm}
\end{figure}


%

$\pname$\footnote{\url{https://github.com/achieve-lab/assertion_data_for_LLM}.} is a comprehensive suite of Verilog design 
and associated formally verified assertions to evaluate the goodness of the COTS LLMs for assertion generation~\cite{pulavarthi2025naacl}. 
$\pname$ consists of designs from OpenCores~\cite{opencores}. ~\Cref{fig:design_deatails} and~\Cref{tab:design_details} show 
representative details of the 
designs. 
Our benchmark consists 
of five ICL examples for 1-shot and 5-shot learning, where each example is a tuple consisting of a Verilog design and its formally verified assertions, generated from $\goldmine$~\cite{goldmine} and HARM~\cite{harm2022tcad}, and verified using Cadence JasperGold~\cite{jaspergold}. The training set comprises fundamental designs such as Arbiter, Half Adder, Full Adder, T-flip-flop, and Full Subtractor. Among these designs, Arbiter and T flip-flops are sequential, while the others are combinational. Our training set assertions contain both overlapped and non-overlapped implication operators. The test design set contains 100 Verilog designs (split in combinational and sequential designs) from OpenCores~\cite{opencores} that are more complex 
than those in the training set, to evaluate 
LLMs' 1-shot and 5-shot learning. The set cover a wide variety of hardware including communication controllers, random number generators (RNG) for security hardware, Floating Point Unit (FPU), state machines, and flow control hardware. The test designs code size varies from 10 lines to 1150 lines (excluding blanks and comments) as measured by cloc~\cite{cloc}.






\begin{table*}
    \centering
    \caption{{\bf Details of a few representative designs in the test set of $\pname$ benchmark}.}
    \label{tab:design_details}    
    \begin{tabular}{l|ll|p{5.2cm}}
        \hline
        {\bf Verilog Design} & {\bf \# of Lines} &  {\bf Design Type} & {\bf Design Functionality} \\ 
        \hline\hline
        ca\_prng & 1144 &  Sequential &  A compact Pattern Generator  \\
        \hline
        cavlc\_read\_total\_coeffs & 1090 & Sequential & 
        Video Encoder 
        for generic audio visual. \\
        \hline
        cavlc\_read\_total\_zeros & 637 & Combinational & Video Encoder for generic audio visual. \\
        \hline
        ge\_1000baseX\_rx &	544 & Sequential & Verilog implementation of Physical Coding Sublayer (PCS) type. \\
        \hline
        MAC\_tx\_Ctrl & 504 & Sequential &   An Ethernet MAC controller. \\ 
        \hline
    \end{tabular}
\end{table*}

\section{Experimental Setup 
}\label{sec:exp_setup}


\begin{figure}
\begin{lstlisting}[language=Verilog, 
                 linewidth=\columnwidth, framexleftmargin=2pt,
                 framexrightmargin=2pt,
                 ]
You are an expert in SystemVerilog Assertions.
Your task is to generate the list of assertions to the given verilog design. An example is shown below. Generate only the list of assertions for the test program with no additional text.
Program 1: module arb2(clk, rst, req1, req2, gnt1, gnt2); input clk, rst; ...
Assertions 1: (state == 1 & req2 == 1) |-> (gnt1 == 0);... 
Test Program: 
module fifo_mem #(parameter DEPTH=8, DATA_WIDTH=8, PTR_WIDTH=3) ( input wclk, w_en, rclk, r_en, input [PTR_WIDTH:0] b_wptr,  ...
Test Assertions:
\end{lstlisting}
\vspace{-2mm}
\caption{{\bf An example of the prompt for 1-shot learning}~\cite{pulavarthi2025naacl}. The example consists of a tuple, a Verilog 
design ({\tt Program 1}) and a set of formally verified assertions for the design ({\tt Assertions 1}). The {\tt Test Program} is the Verilog 
design for which we generate assertions using the trained LLM.}
\label{fig:prompt_1shot}
\vspace{-5mm}
\end{figure}



\noindent {\bf Evaluation Protocol}: \Cref{fig:eval_framework} shows our evaluation framework. To evaluate the effectiveness of the different LLMs, our $k$-shot ICL 
consists of 1-shot and 5-shot in-context examples (ICE) (\circled{1} in~\Cref{fig:eval_framework}). Each 
ICE is a tuple $\langle \mathcal{D}, \mathcal{A} \rangle$, where $\mathcal{D}$ is a Verilog 
source code and $\mathcal{A}$ is a set of formally verified assertions containing a minimum of two (2) and a maximum of 10 assertions with an 
average of 
4.8 assertions per source code.
We use a prompt as shown in~\Cref{fig:prompt_1shot} consisting of four parts -- (i) an English language description of the task in hand, (ii) an example 
Verilog design with newlines and comments removed, (iii) an example assertion in SVA format, and (iv) a test Verilog design with new lines and comments removed. Followed by training, we provide each trained model with 100 test Verilog programs to infer assertions (\circled{2} in~\Cref{fig:eval_framework}). In our experiments, we have found all of the LLM models generate syntactically erroneous assertions, \ie, each LLM fails to learn the SVA syntax from the training examples. Consequently, we use a syntax corrector (\circled{3} in~\Cref{fig:eval_framework}) using $\gptt$ and feed the output of the syntax corrector to Cadence JasperGold FPV engine to evaluate the quality of the generated assertions. Note any other FPV engine compatible with SVA will work as well.


\smallskip

\noindent {\bf ICL Compute Platform}: We use UIUC (University of Illinois Urbana-Champaign) NCSA's (National Center for Supercomputing Applications) 
Delta Cluster~\cite{ncsadelta} 
to run our experiments. 
We use GPU nodes containing 1-way, 4-way, and 8-way NVIDIA A40 (with 48GB GDDR6) and A100 (with 40GB SXM ) GPUs to perform $k$-shot learning. 

\smallskip

\noindent {\bf Pre-trained Models and EDA Tools}: We use pre-trained LLMs from the HuggingFace~\cite{huggingface} 
for 
evaluation and 
Cadence JasperGold version 2022.06p002 
to formally verify the assertions generated from the test Verilog designs. We use two SOTA 
tools $\goldmine$~\cite{goldmine, pal2020tcad} and HARM~\cite{harm2022tcad} to generate assertions for 
Verilog designs in the ICE. 
Below, we summarize the 
COTS LLMs that we evaluate using 
$\pname$. 

\begin{enumerate}[leftmargin=*, parsep=0cm, itemindent=1.2em, itemsep=0.2em, topsep=0.1em]
    \item {\bf $\gptt$} 
    is a commercial LLM 
built using the GPT architecture~\cite{chatgpt35}. It is part of OpenAI's GPT (Generative Pre-trained Transformer) 
series of models designed to understand and generate text based on the input it receives. 




\item {\bf $\gptf$} 
(`o' for ``omni'') is the newest model of OpenAI's GPT, which accepts any combination of input audio, image, video, and text and responds with an output consisting of image, audio, video, and text~\cite{openai2024gpt4}. 
With larger training data, increased model size, and faster response than 
other GPT models, $\gptf$ is a unified 
model for 
text, vision, and audio. 



\item {\bf $\cllama$} 
is a collection of 
generative text models developed by Meta~\cite{codellama2} with parameters ranging 
from 7B to 70B. 
The model accepts only text as input and output. 
It is an auto-regressive language model. 
The context window length for $\cllama$ is 4096. 
The large 70B 
model uses Grouped-Query Attention for improved inference scalability. 


\item {\bf  $\llama$} 
is available in two parameter sizes -- 8B and 70B. The context window length for $\llama$ is 8192 and 
is pre-trained with 15 Trillion tokens of publicly available data~\cite{llama3}. $\llama$ excels at 
translation, contextual understanding, and dialogue generation. It has enhanced capabilities such as code generation, reasoning, and following instructions.

\end{enumerate}

\smallskip

\noindent {\bf ICL Hyperparameters}: For all LLMs 
under consideration, the hyperparameters have been 
set 
to their default values. 
Specifically, the maximum output tokens  is set to 
1024, employing a greedy decoding strategy and maintaining a \textit{temperature} of 1.0 (most creative), \textit{top\_p} of 0.95. The \textit{random seed} is set to 
50.

\smallskip 

\noindent {\bf Metrics}: We evaluate the generated assertions 
from the test designs 
using 
following 
metrics for each $k$-shot ICL per LLM. 

\begin{enumerate}[leftmargin=*, parsep=0cm, itemindent=1em, itemsep=0.2em, topsep=0.1em]
    \item {\bf Pass} 
    quantifies the fraction of generated assertions that FPV engine attests as 
    valid for the design. This includes the Vacuous and the Pass cases from~\Cref{tab:assertion_eval}.
    \item {\bf Fail} 
    quantifies the fraction of generated assertions that FPV engine attests as 
    wrong for the design and generates a counterexample trace. This includes the Fail case from~\Cref{tab:assertion_eval}.
    \item {\bf Error}: It quantifies the fraction of generated assertions for which the FPV engine identifies 
    one or more syntactic errors in the assertions even after syntax correction by the $\gptt$.
\end{enumerate}

%





\begin{figure*}
    \centering
    \begin{subfigure}[b]{0.2\textwidth}
        \includegraphics[scale=0.2]{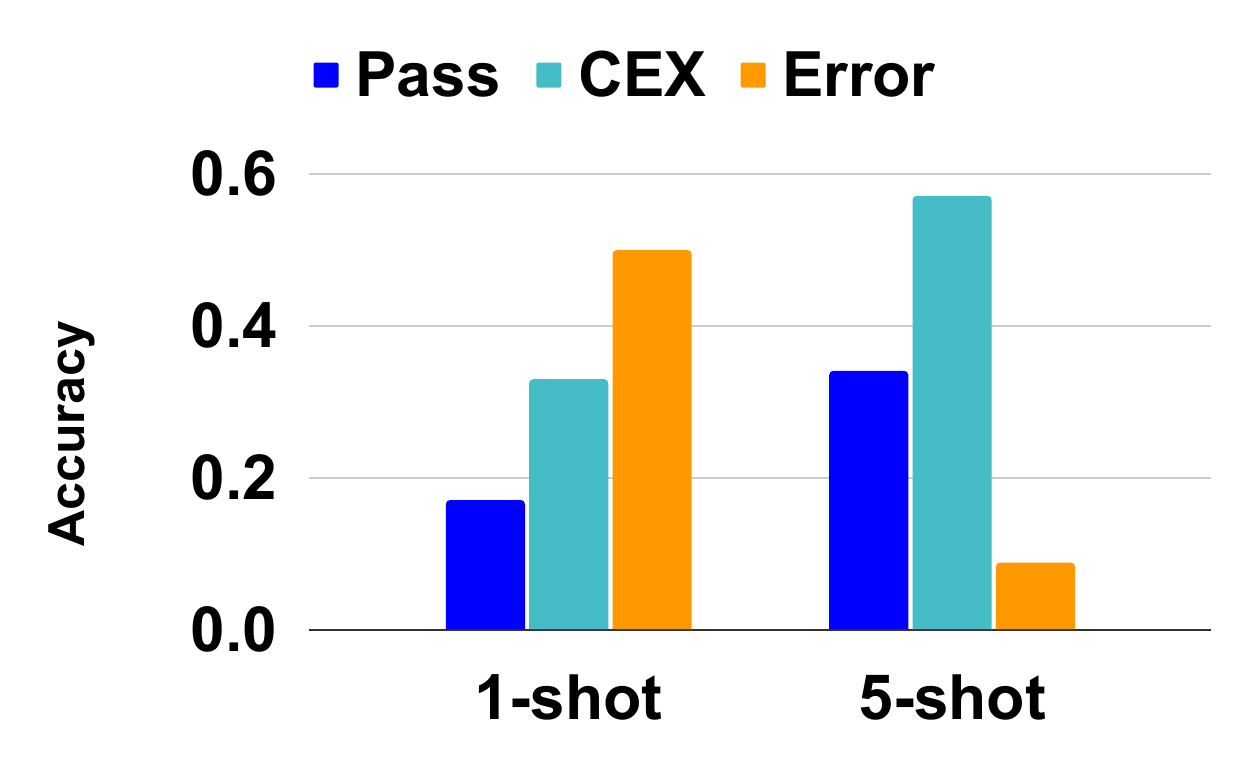}
        \vspace{-5mm}
        \caption{
        \label{fig:gpt35}}
    \end{subfigure}
    \hfill
    \begin{subfigure}[b]{0.2\textwidth}
        \includegraphics[scale=0.2]{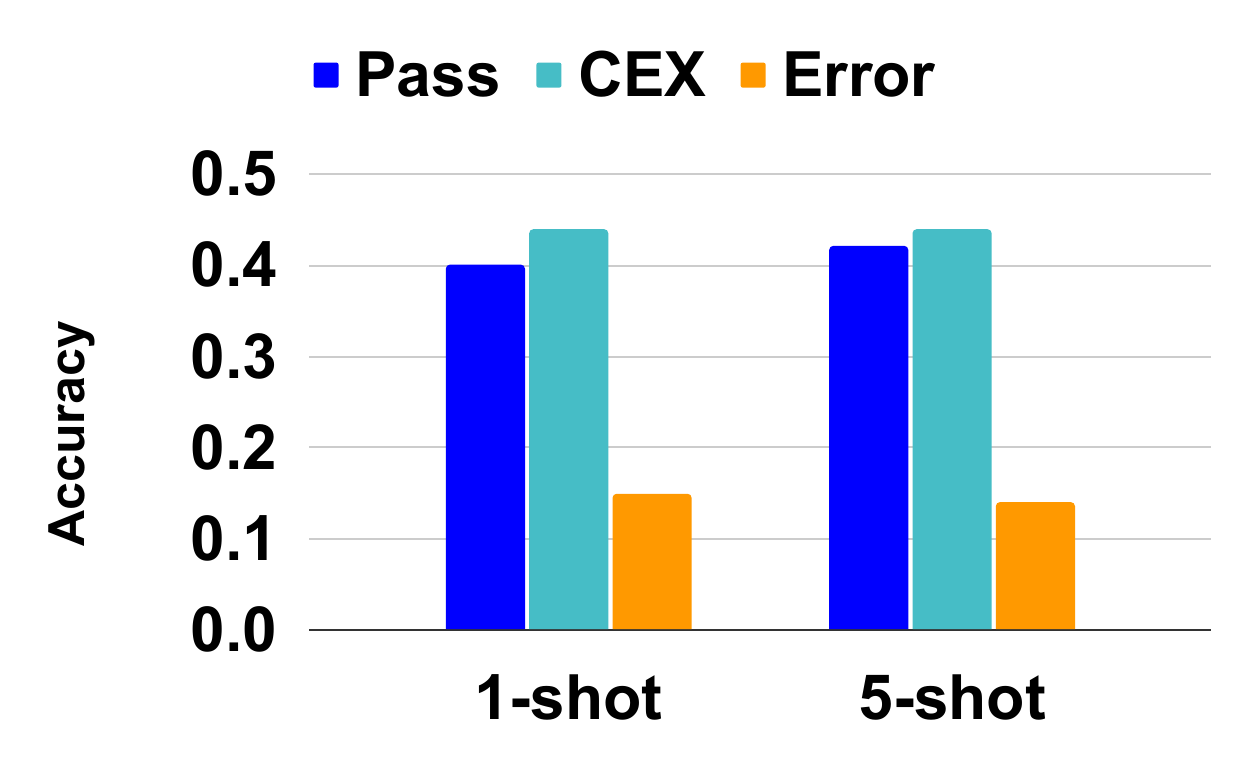}
        \vspace{-5mm}
        \caption{
        \label{fig:gpt4o}}
    \end{subfigure}
    \hfill
    \begin{subfigure}[b]{0.2\textwidth}
        \includegraphics[scale=0.2]{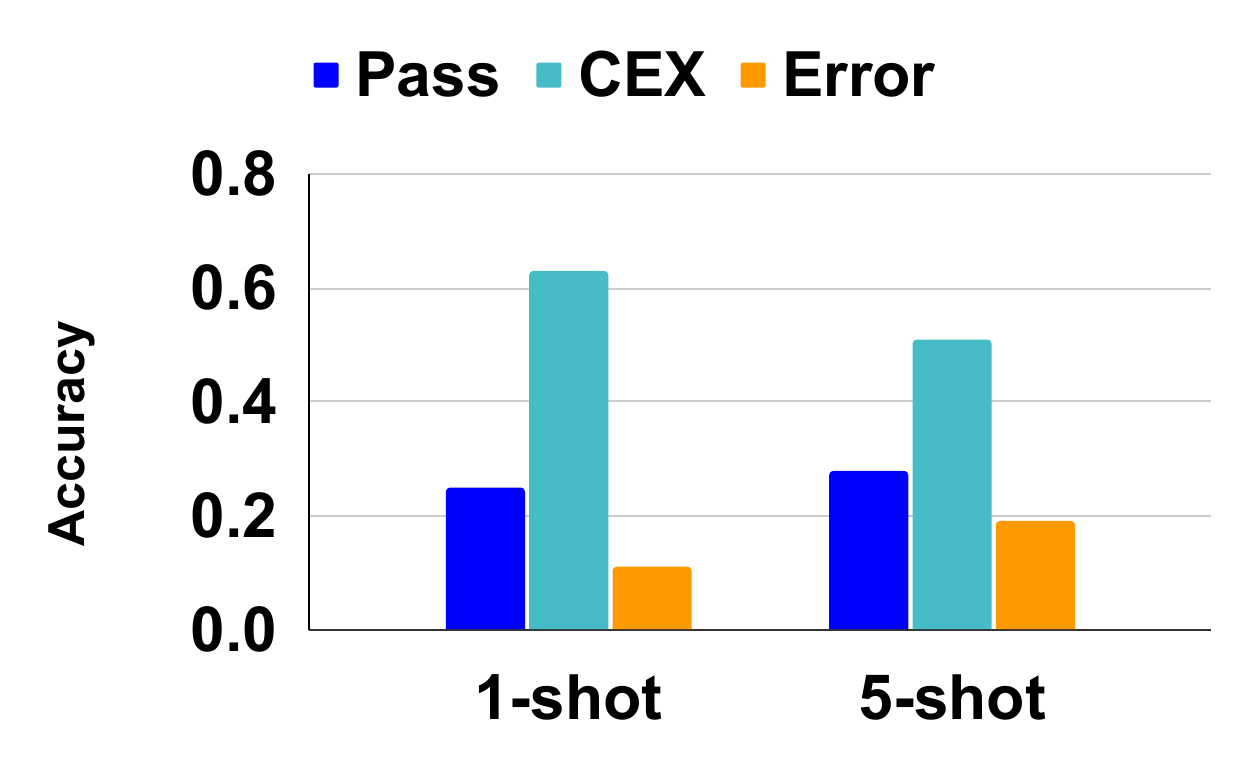}
        \vspace{-5mm}
        \caption{
        \label{fig:codellama2}}
    \end{subfigure}
    \begin{subfigure}[b]{0.2\textwidth}
        \includegraphics[scale=0.2]{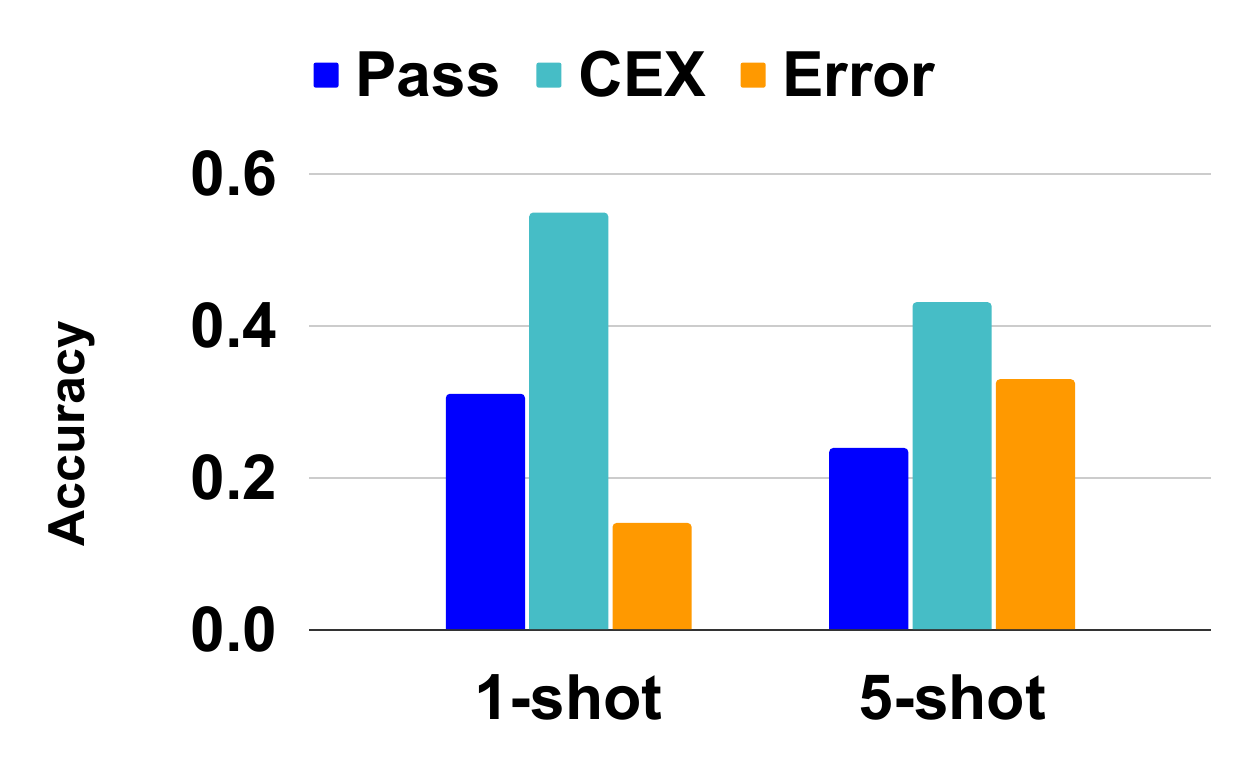}
        \vspace{-5mm}
        \caption{
        \label{fig:llama3}}
    \end{subfigure}
    \hfill
    \vspace{-2mm}
    \caption{{\bf Comparison of accuracy of generated assertions}. (\subref{fig:gpt35}) Assertion accuracy comparison for $\gptt$. (\subref{fig:gpt4o}) Assertion accuracy comparison for $\gptf$. (\subref{fig:codellama2}) Assertion accuracy comparison for $\cllama$. (\subref{fig:llama3}) Assertion accuracy comparison for $\llama$. (\subref{fig:1_shot_acc}) $k = 1$-shot assertion accuracy. (\subref{fig:5_shot_acc}) $k = 5$-shot assertion accuracy. {\bf CEX}: Counter Example trace.
    }
    \label{fig:imp_between_shots}
    \vspace{-5mm}
\end{figure*}

\section{Observations and Insights from $\pname$ 
}\label{sec:exp_res}

\begin{figure}
    \centering
    \begin{subfigure}[b]{0.22\textwidth}
        \centering
        \includegraphics[scale=0.165]{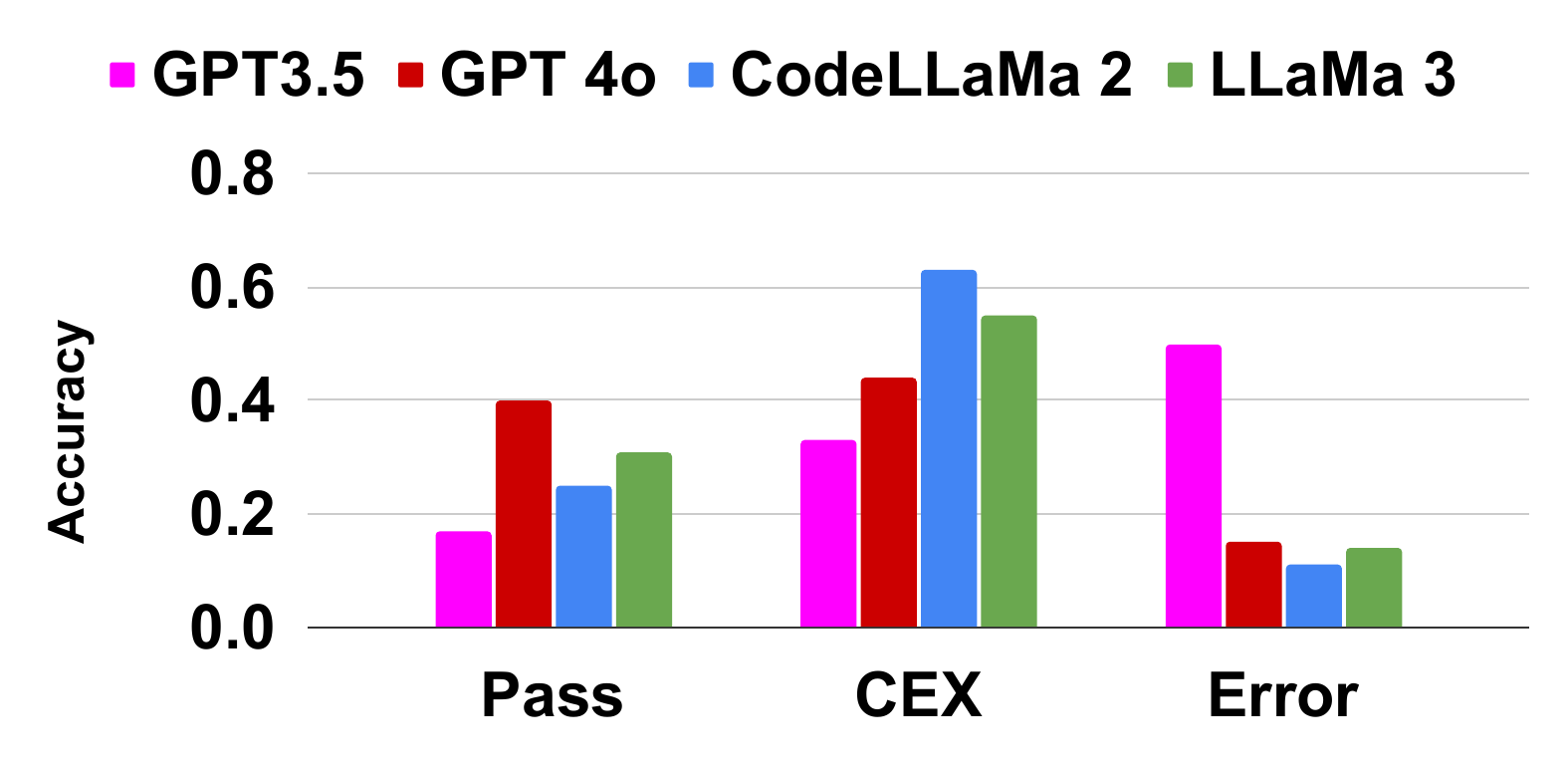}
        \caption{1-shot.\label{fig:1_shot_acc}}
    \end{subfigure}
    \hspace{3mm}
    \begin{subfigure}[b]{0.22\textwidth}
        \centering
        \includegraphics[scale=0.165]{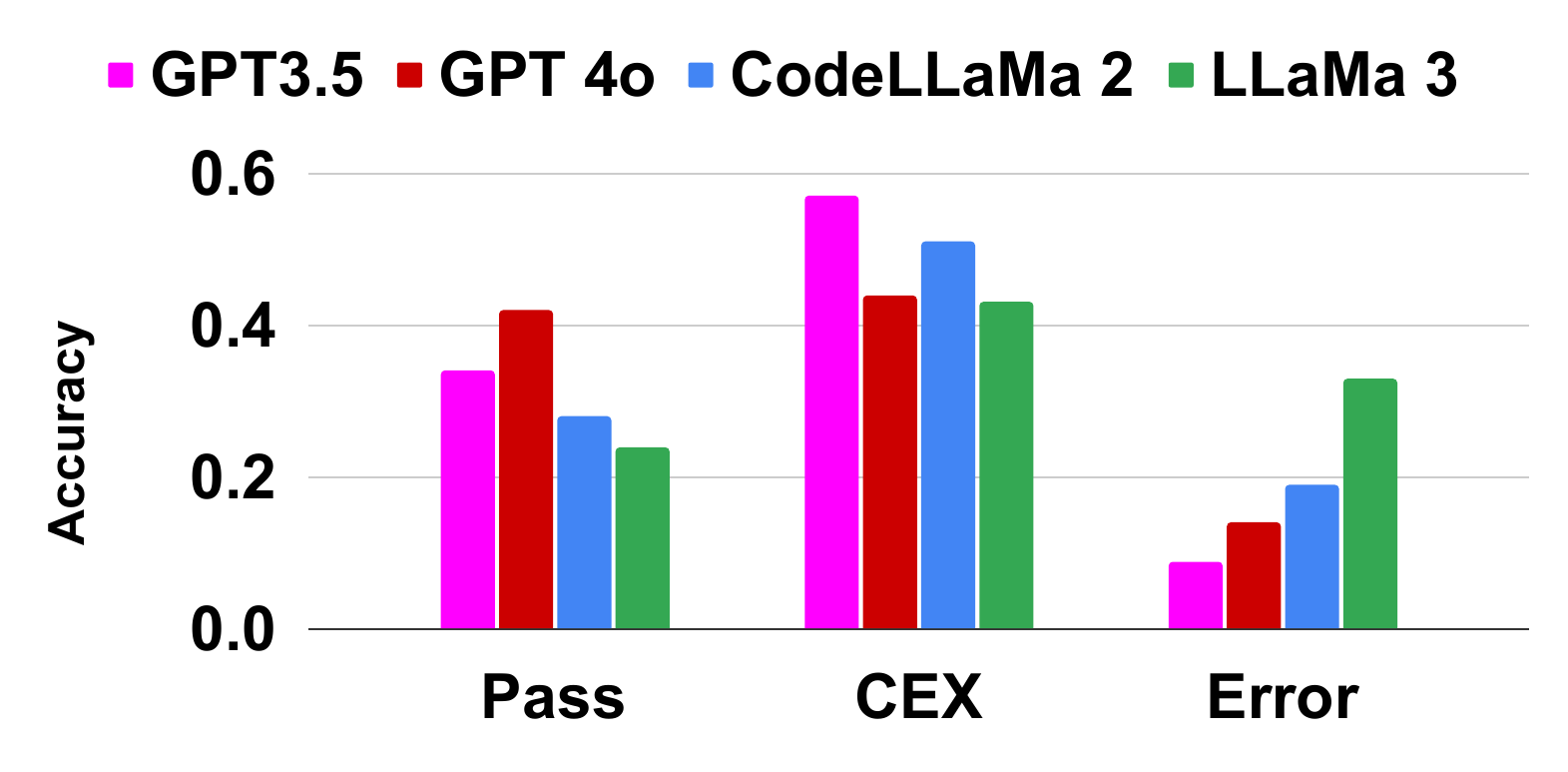}
        \caption{5-shot.\label{fig:5_shot_acc}}
    \end{subfigure}   
    \caption{{\bf Comparison of accuracy of generated assertions in terms of passing, failing, (generating a counter example), and syntactically wrong assertions between different LLMs per $k$-shot learning where $k = 1$ and $k = 5$}. 
    }
    \label{fig:accuracy_across_LLMs}
    \vspace{-6mm}
\end{figure}

We depict our observations and insights~\cite{pulavarthi2025naacl} 
in~\Cref{fig:imp_between_shots} and~\Cref{fig:accuracy_across_LLMs} and discuss them below. 

\smallskip

\noindent \bem{Observation 1}: {\bf Most LLMs generate valid assertions with an increasing number of ICL examples}. For the assertion generation task, all LLMs progressively generate more valid assertions when the number of ICL examples is increased as seen in~\Cref{fig:imp_between_shots}. $\gptt$, $\gptf$, and $\cllama$ show on average an improvement of 2$\times$, 1.2$\times$, and 1.12$\times$ for valid assertion generation, respectively, when moved from 1-shot learning to 5-shot learning. However, the $\llama$ model loses accuracy from 31\% to 24\% on the same dataset. Our in-depth analysis shows in many cases, $\llama$ either fails to generate assertions or generates syntactically wrong assertions (which even a syntax corrector fails to correct) or tries to generate codes in a new programming language (such as Java). \bem{This experiment shows that there is a considerable scope for 
improving the $\llama$ model for this task, likely via fine-tuning the pre-trained $\llama$ model}. 

\smallskip

\noindent \bem{Observation 2}:  {\bf An enhanced LLM does not necessarily ensure a better semantic or syntactic understanding}. In~\Cref{fig:imp_between_shots}, we 
do not see a 
correlation between the sophistication (in terms of the number of model parameters) of the LLMs and their ability to generate 
good assertions. For $\gptt$ (\cf,~\Cref{fig:gpt35}), with an increase in the number of ICL examples, the LLM was able to produce more syntactically correct assertions, however, after such corrections, the majority of assertions (on average up to 24\%) generated a CEX when verified with JasperGold. For $\gptf$, the results were more consistent in terms of syntactically (in)correct assertions for 
both 
1-shot and  5-shot learning (\cf,~\Cref{fig:gpt4o}). For $\cllama$ and $\llama$, with an increase in the number of ICL examples, the number of failed assertions decreased (on average up to 12\% for $\cllama$ and $\llama$, \cf,~\Cref{fig:codellama2} and~\Cref{fig:llama3}), however, both models generated more syntactically wrong assertions (on average up to 19\% more for $\llama$). This observation is perplexing as one would expect with more number of parameters, $\llama$ would be able to learn better to predict syntactically correct assertions. Our in-depth analysis shows that with a 1-shot, the variations in types of assertions in examples were limited. Consequently, $\llama$ learned the assertion syntax. However, in 5-shot learning, we have more variations in the 
which made $\llama$'s learning task difficult. \bem{This experiment suggests that increasing the ICL examples alone will not necessarily improve LLM's consistency in generating syntactically and semantically 
correct assertions}. 

\smallskip

\noindent \bem{Observation 3}: {\bf $\gptf$ is relatively more consistent for assertion generation task}. In~\Cref{fig:accuracy_across_LLMs}, we compare different LLMs in terms of generating valid assertions for 1-shot and 5-shot learning. Our experiments show that $\gptf$ is consistently superior in generating valid assertions for both 1-shot and 5-shot learning and generates, on average, up to 15.6\% more valid assertions as compared to other LLMs. This trend remains valid with respect to assertions generating CEX and syntactically wrong assertions, \ie, $\gptf$ produced less CEX generating assertions and syntactically incorrect assertions as compared to other LLMs. \bem{This experiment shows 
that $\gptf$ performs relatively better than other LLMs}. 

\smallskip

\noindent \bem{Observation 4}: {\bf All LLMs need considerable improvement for assertion generation task}: In-depth analysis of~\Cref{fig:imp_between_shots} and~\Cref{fig:accuracy_across_LLMs} show that none of the LLMs can generate valid assertions with an average of no more than 44\% accuracy whereas up to 63\% generated assertions produces CEX and on average up to 33\% of generated assertions are syntactically wrong. Clearly, for LLMs to be of practical usage for any realistic industrial-scale design, considerable improvement needs to be made. Specifically, the LLMs need to capture the semantic meaning of the hardware description languages, \eg, Verilog, 
for generating higher fraction of valid assertions automatically without iterative human prompting. Our 
prior work~\cite{goldmine, pal2020tcad} shows 
that such critical insights are not directly available from the raw design source code and need auxiliary artifacts, such as Control-Data Flow Graph (CDFG), Variable Dependency Graph (VDG), Cone of Influence (COI), etc. \bem{Future research in applying LLMs for assertion generation should consider such auxiliary artifacts to develop 
assertion-specific LLMs}.


Evaluation of the four COTS LLMs using $\pname$ shows that no LLM consistently outperforms other LLMs. Our analysis emphasizes that there is a considerable scope to enhance LLMs for assertion generation. \bem{There are two different ways} -- (i) enhance ICL with diverse 
ICL examples or (ii) develop 
an LLM specifically for assertion generation. Using this insight, we develop $\allm$ as detailed in~\Cref{sec:allm}.

\begin{figure}
    \centering
    \includegraphics[scale=0.3]{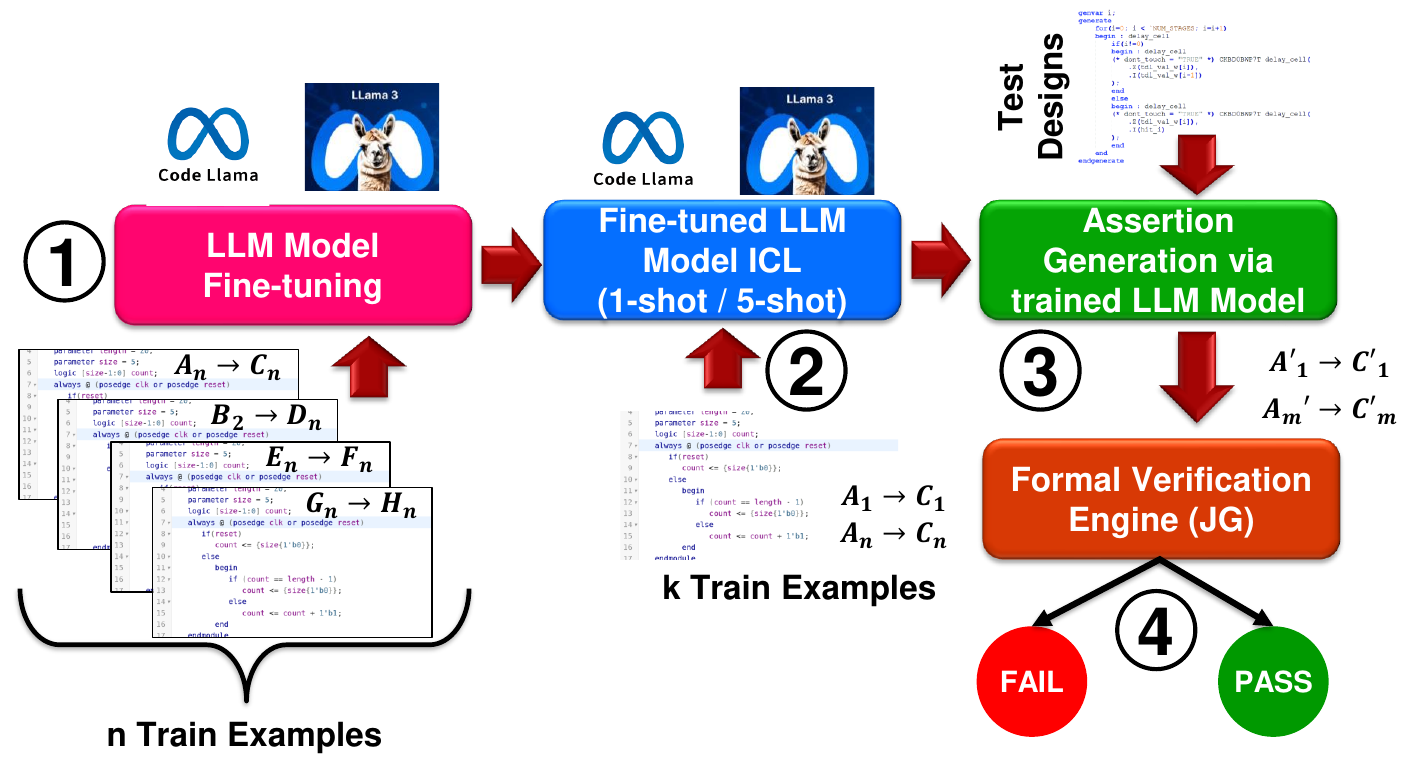}
    \caption{{\bf Our evaluation framework for $\allm$}.}
    \label{fig:assertionllm}
    \vspace{-5mm}
\end{figure}

\section{$\allm$: LLM for Assertion Generation}\label{sec:allm}

$\allm$ is an LLM specifically developed for the task of assertion generation. This is inline with the recent findings in other domains, \eg, financial analysis, where a task-specific LLM has excelled for downstream tasks as compared to a foundational LLM. We use two different LLM foundational models -- (i) $\cllama$ and (ii) $\llama$ and finetune each of them using a large amount of data where each data point consists of a Verilog design and its formally verified assertions.

~\Cref{fig:assertionllm} shows the end-to-end flow to finetune LLM for assertion generation and how we use the finetuned LLM to evaluate its goodness for assertion generation. Compared to~\Cref{fig:eval_framework}, we have removed the syntax corrector block (\circled{3} in~\Cref{fig:eval_framework}), and instead of using foundational LLM (\circled{1} in~\Cref{fig:eval_framework}), we are using the fined-tuned LLM model for ICL (\circled{2} and \circled{3} in~\Cref{fig:assertionllm}).

We use the same compute platform and and hyperparameters as detailed in~\Cref{sec:exp_setup} for finetuning. Additionally, we use 20 epochs to finetune each foundational LLM. We split the $\pname$ and use 75\% of data for training and the remaining 25\% for testing the goodness of the tuned LLM.

\section{Experimental Results on $\allm$}\label{sec:exp_res_allm}

\begin{figure}
    \centering
    \begin{subfigure}[b]{0.22\textwidth}
        \centering
        \includegraphics[scale=0.21]{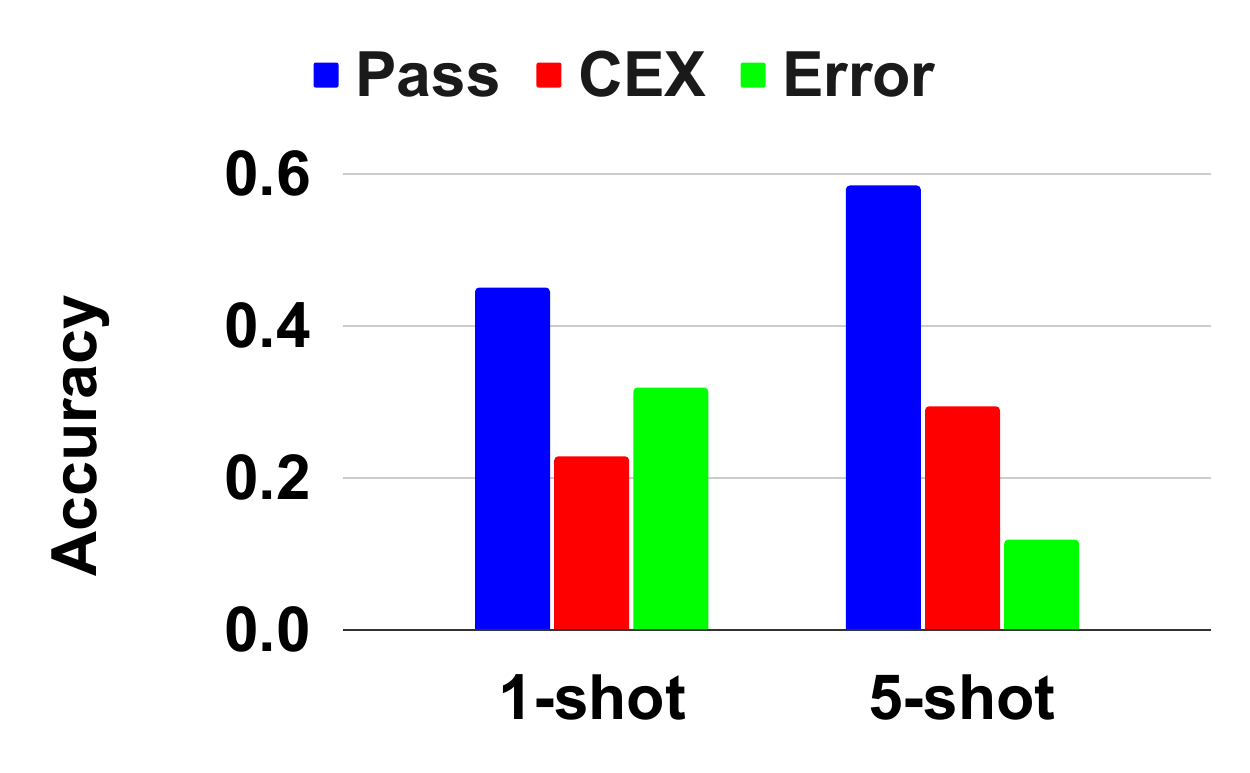}
        \caption{\label{fig:codellama_finetune}}
    \end{subfigure}
    \hspace{3mm}
    \begin{subfigure}[b]{0.22\textwidth}
        \centering
        \includegraphics[scale=0.21]{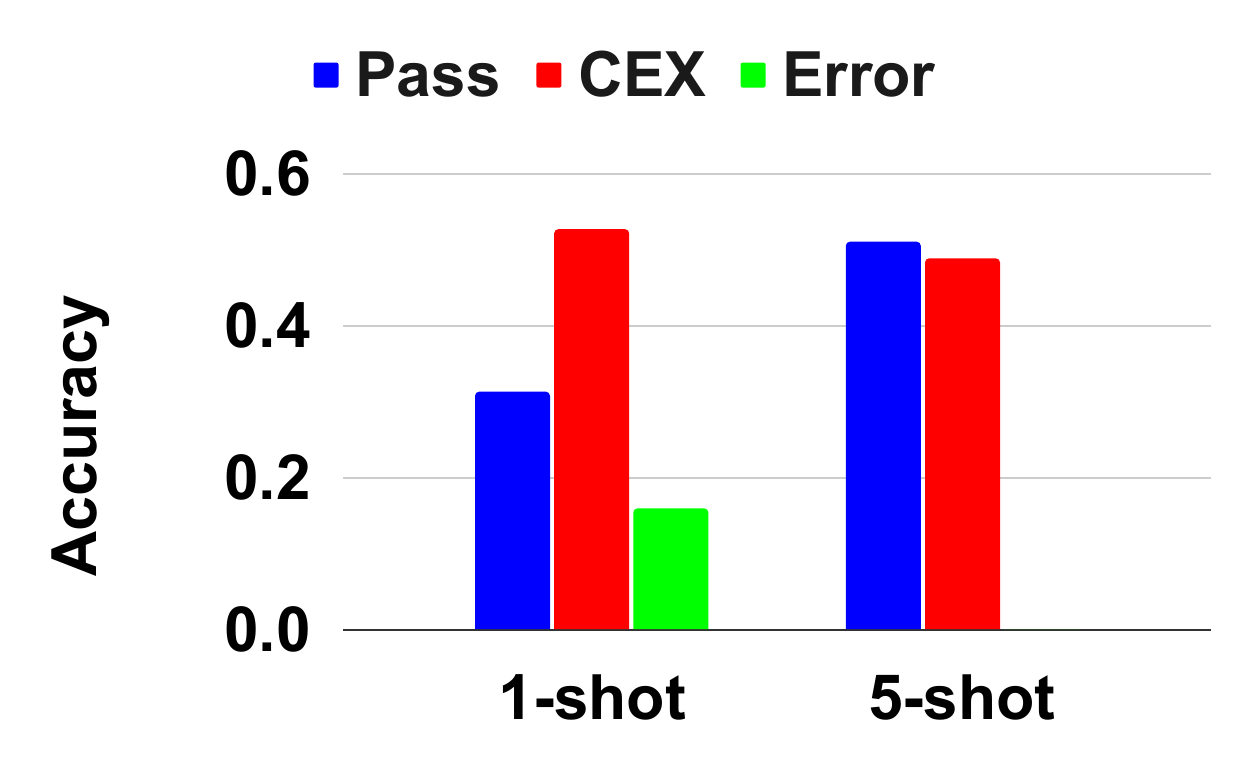}
        \caption{\label{fig:llama_finetune}}
    \end{subfigure}   
    \caption{{\bf Comparison of accuracy of generated assertions}. (\subref{fig:codellama_finetune}) Assertion accuracy comparison for finetuned $\cllama$. (\subref{fig:llama_finetune}) Assertion accuracy comparison for finetuned $\llama$.  {\bf CEX}: Counter Example trace. 
    }
    \label{fig:accuracy_across_finetuned_LLMs}
    \vspace{-6mm}
\end{figure}

~\Cref{fig:accuracy_across_finetuned_LLMs} shows our experimental results on assertion generation task using finetuned $\cllama$ and $\llama$. We compare these results to that of results in~\Cref{fig:imp_between_shots}.

\smallskip

\noindent \bem{Observation 5}: {\bf Finetuning LLMs considerably improves fraction of correct assertions}: We observe that the finetuned $\cllama$ increased proven assertions by 29\% and 38\% for 1-shot and 5-shot ICL, decreased assertions generating CEX by 48\% and 33\% for 1-shot and 5-shot ICL, respectively (\cf,~\Cref{fig:codellama2} and~\Cref{fig:codellama_finetune}). With respect to $\llama$, fine tuning has increased proven assertions by 24\% for 5-shot learning (\cf,~\Cref{fig:llama3} and~\Cref{fig:llama_finetune}). However, for 1-shot ICL, the fraction of proven assertions has decreased by 4.7\% and has increased the fraction of assertions generating CEX by 5.4\% and 12\% for 1-shot and 5-shot ICL, respectively. Further analysis shows that as the foundational $\cllama$ was 
trained on large corpora of codes, it learned assertion syntax and semantics better during finetuning. In contrast, foundational $\llama$ training on general text corpora struggles to learn assertion syntax and semantics during finetuning. \bem{This experiment shows that it is crucial to select appropriate foundational LLM and dataset for an effective fine-tuned LLM for assertion generation}.


\smallskip

\noindent \bem{Observation 6}: {\bf Fine-tuning LLMs does not necessarily guarantee syntactic error-free assertions}: Fine-tuning LLMs does not necessarily nullify the fraction of syntactically erroneous assertions.~\Cref{fig:accuracy_across_finetuned_LLMs} shows that both finetuned $\cllama$ and $\llama$ generate a considerable fraction (upto 38\%) of syntactically erroneous assertions. \bem{In order to reduce the fraction of erroneous assertions, we speculate that we will require a more comprehensive dataset with a sufficient number of examples for the diverse syntax of assertions}.

\section{Related Work 
}\label{sec:rel_work}



Automatic generation of assertions in 
hardware has been an active area of research for the past decade. 
IODINE is one of the earliest works for hardware assertion generation by analyzing dynamic program behavior with respect to a set of standard property templates~\cite{iodine2005}. Prior works have used static analysis~\cite{pinter2005hase, hekmatpour2005}, dynamic simulation execution data~\cite{inferno2009tcad, chang2015aspdac, chung2011iccd}, and  data-driven statistical analysis guided by the lightweight static analysis of design source code~\cite{goldmine, hertz2013tacd} 
for assertion generation. 
Following $\goldmine$, researchers have developed a wide variety of assertion generation techniques
targeting hardware design functionality~\cite{hertz2019aspdac, pal2020tcad, iman2024artmine, germiniani2022vlsi, iman2022dsd, harm2022tcad} and hardware design security~\cite{calvin2021ashes, hasini2023jetc}, and 
to evaluate the quality of numerous assertions that automatic methods generate 
to aid the downstream verification tasks~\cite{avinash2024host, pal2020tcad, iman2022dsd}. \bem{However, all these works suffer from following shortcomings} -- they (i) 
do not scale for industrial-scale designs, (ii) 
require a massive amount of 
trace data to 
generate assertions, 
(iii) 
generate numerous redundant 
design properties 
without any explanation on their usability for downstream verification tasks, (iv) 
fail to generalize the properties beyond what is seen in the trace, and (v) 
encompass an extremely small subset of SVA, 
limiting the expressibility and richness of the generated assertions. 


Recently, massive success of LLMs, \eg, GPT~\cite{openai2024gpt4}, LLaMa~\cite{touvron2023llama}, 
Gemini~\cite{gemini2024google}, in diverse scientific, engineering, and medical applications have led researchers to investigate application of 
LLMs for hardware property generation 
~\cite{liu2023verilogeval, orenesvera2023using, kande2023llmassisted, fang2024assertllm, mali2024chiraag, stutton2023icml}. \bem{However, almost all recent works on property generation using LLMs suffer from the following shortcomings} -- they (i) 
require considerable 
\fixme{human efforts and  deep 
understanding of the target hardware designs to devise {\em hand-crafted prompts}} to generate and refine 
hardware properties,  
(ii) 
do not generalize the assertions, and (iii) 
do not consider execution traces, risking potentially missing subtle incorrect design behaviours or security vulnerabilities 
that are not obvious in the design source code. In fact, there is a lack of 
a systematic study comparing the effectiveness of different commercial and open-source LLMs 
in generating valid assertions from 
hardware design source code. $\pname$ aims to fill in the gap and provides novel insights for future research on LLMs for assertion generation.



\section{Limitations}\label{sec:limitations}

\begin{itemize}[leftmargin=*, parsep=0cm, itemindent=1.2em, itemsep=0.2em, topsep=0.1em]
    \item {\bf Dataset}: In this study, we focus on Verilog designs, given its predominance in hardware design language. Moving forward, it will be intriguing to develop benchmarks for assertions in other HDLs, 
    \eg, VHDL, SystemC, to 
    expand 
    the scope of our analysis to 
    broader 
    design paradigms.
    \item {\bf Modeling}: In this paper, we assessed the assertion generation capabilities of $k$-shot and finetuned SOTA LLMs. There is a considerable scope for improvement in terms of assertion quality and correctness. Future work should focus on modeling to capture design coverage of generated assertions and quantify their goodness in terms of captured design behavior. 
    \item {\bf Evaluation}: In future work, it will be valuable to conduct a more detailed evaluation of model errors to better understand the specific limitations of each LLM for assertion generation.
\end{itemize}

\section{Conclusion and Future Work 
}\label{sec:conclusion}

This work introduces $\pname$ to evaluate the current and future commercial and open-source LLMs for the assertion generation task and $\allm$ to fully automate assertion generation using generative AI without the designer's iterative intervention. Although there is no LLM that consistently outperforms other LLMs, we notice several promising trends and research directions such as (i) to quantify the goodness of assertion in terms of captured design behavior, (ii) to quantify the design coverage of the assertions, (iii) to model and capture likely design security vulnerabilities as assertions, and (iv) going beyond temporal/sequential assertions to generate assertions encompassing richer set of SVA, to enhance the practical applicability of LLMs for assertion generation task for industrial-scale designs. Pursuing these directions will further accelerate SoC and hardware design verification.


\clearpage
\newpage


\end{document}